\documentclass[10pt,twocolumn,letterpaper]{article}

\usepackage{cvpr}
\usepackage{times}
\usepackage{epsfig}
\usepackage{graphicx}
\usepackage{amsmath}
\usepackage{amssymb}
\usepackage{multirow}
\usepackage{caption}
\usepackage{subfigure}
\usepackage{authblk}

\usepackage[breaklinks=true,bookmarks=false]{hyperref}

\cvprfinalcopy 

\def\cvprPaperID{****} 

\begin{document}
\language0
\lefthyphenmin=2
\righthyphenmin=3

\title{Object as Hotspots: An Anchor-Free 3D Object Detection Approach \\ via Firing of Hotspots}

\author{Qi Chen$^{1, 2}$\thanks{Work done while interning at Samsung.}\ , Lin Sun$^{1}$\thanks{Corresponding author: lin1.sun@samsung.com}, Zhixin Wang$^3$, Kui Jia$^3$, and Alan Yuille$^1$\\
$^1$Samsung Semiconductor, Inc.\\
$^2$Johns Hopkins University\\ 
$^3$South China University of Technology\\
}
\maketitle

\begin{abstract}
Accurate 3D object detection in LiDAR based point clouds suffers from the challenges of data sparsity and irregularities. Existing methods strive to organize the points regularly, e.g. voxelize, pass them through a designed 2D/3D neural network, and then define object-level anchors that predict offsets of 3D bounding boxes using collective evidences from all the points on the objects of interest. Contrary to the state-of-the-art anchor-based methods, based on the very nature of data sparsity, we observe that even points on an individual object part are informative about semantic information of the object. We thus argue in this paper for an approach opposite to existing methods using object-level anchors. Inspired by compositional models, which represent an object as parts and their spatial relations, we propose to represent an object as composition of its interior non-empty voxels, termed hotspots, and the spatial relations of hotspots. This gives rise to the representation of Object as Hotspots (OHS). Based on OHS, we further propose an anchor-free detection head with a novel ground truth assignment strategy that deals with inter-object point-sparsity imbalance to prevent the network from biasing towards objects with more points. Experimental results show that our proposed method works remarkably well on objects with a small number of points. Notably, our approach ranked $1^{st}$ on KITTI 3D Detection Benchmark for cyclist and pedestrian detection, and achieved state-of-the-art performance on NuScenes 3D Detection Benchmark.

\end{abstract}

\section{Introduction}
Great success has been witnessed in 2D detection recently thanks to the evolution of CNNs. However, extending 2D detection methods to LiDAR based 3D detection is not trivial because point clouds have very different properties from those of RGB images. Point clouds are irregular, so \cite{zhou2018voxelnet,maturana2015voxnet,yang2018pixor} have converted the point clouds to regular grids by subdividing points into voxels and process them using 2D/3D CNNs. Another unique property and challenge of LiDAR point clouds is the sparseness. LiDAR points lie on the objects' surfaces and meanwhile due to occlusion, self-occlusion, reflection or bad weather conditions, very limited quantity of points can be captured by LiDAR. 

Inspired by compositional part-based models \cite{jin2006context,zhu2008,fidler2014,dai2014unsupervised,kortylewski2017greedy}, which have shown robustness when classifying partially occluded 2D objects and for detecting partially occluded object parts \cite{zhang2018deepvoting}, we propose to detect objects in LiDAR point clouds by representing them as composition of their interior non-empty voxels. We define the non-empty voxles which contain points within the objects as \textbf{spots}. Furthermore, to encourage the most discriminative features to be learned, we select a small subset of spots in each object as \textbf{hotspots}, thus introducing the concept of hotspots. The selection criteria are elaborated in Sec. \ref{hotspot}. Technically, during training, hotspots are spots assigned with positive labels; during inference hotspots are activated by the network with high confidences.  

Compositional models represent objects in terms of object parts and their corresponding spatial relations. For example, it can not be an actual dog if a dog's tail is found on the head of the dog. We observe the ground truth box implicitly provides relative spatial information between hotspots and therefore propose a \textbf{spatial relation encoding} to reinforce the inherent spatial relations between hotspots.

We further realize that our hotspot selection can address an \textbf{inter-object point-sparsity imbalance} issue caused by different object sizes, different distances to the sensor, different occlusion/truncation levels, and reflective surfaces etc. A large number of points are captured on large objects or nearby objects to the sensor while much fewer points are collected for small objects and occluded ones. In the KITTI training dataset, the number of points in annotated bounding boxes ranges from $4874$ to $1$. We categorize this issue as feature imbalance: objects with more points tend to have rich and redundant features for predicting semantic classes and localization while those with few points have few features to learn from. 

The concept of hotspots along with their spatial relations gives rise to a novel representation of \textbf{Object as Hotspots (OHS)}. Based on OHS, we design an OHS detection head with a hotspot assignment strategy that deals with inter-object point-sparsity imbalance by selecting a limited number of hotspots and balancing positive examples in different objects. This strategy encourages the network to learn from limited but the most discriminative features from each object and prevents a bias towards objects with more points.

Our concept of OHS is more compatible with anchor-free detectors. Anchor-based detectors assign ground truth to anchors which match the ground truth bounding boxes with IoUs above certain thresholds. This strategy is object-holistic and cannot discriminate different parts of the objects while anchor-free detectors usually predict heatmaps and assign ground truth to individual points inside objects. However, it’s nontrivial to design an anchor-free detector. Without the help of human-defined anchor sizes, bounding box regression becomes difficult. We identify the challenge as \textbf{regression target imbalance} due to scale variance and therefore adopt soft $argmin$ from stereo vision \cite{kendall2017end} to regress bounding boxes. We show the effectiveness of soft $argmin$ in handling regression target imbalance in our algorithm.

The main contributions of proposed method can be summarized as follows:
\begin{itemize}
\item We propose a novel representation, termed Object as HotSpots (OHS) to compositionally model objects from LiDAR point clouds as hotspots with spatial relations between them.

\item We propose a unique hotspot assignment strategy to address inter-object point-sparsity imbalance and adopt soft $argmin$ to address the regression target imbalance in anchor-free detectors.

\item Our approach shows robust performance for objects with very few points. The proposed method sets the new state-of-the-art on Nuscene dataset and KITTI test dataset for cyclist and pedestrian detection. Our approach achieves real-time speed with $25$ FPS on KITTI dataset.

\end{itemize}

\section{Related Work}
\subsubsection{Anchor-Free Detectors for RGB Images} Anchor-free detectors for RGB images represent objects as points. Our concept of object as hotspots is closely related to this spirit. ExtremeNet \cite{zhou2019bottom} generates the bounding boxes by detecting top-most, left-most, bottom-most, right-most, and center points of the objects. CornerNet \cite{law2018cornernet} detects a pair of corners as keypoints to form the bounding boxes. Zhou et al \cite{zhou2019objects} focuses on box centers, while CenterNet \cite{duan2019centernet} regards both box centers and corners as keypoints. FCOS \cite{tian2019fcos} and FSAF\cite{zhu2019feature} detect objects by dense points inside the bounding boxes. The difference between these detectors and our OHS is, ours also takes advantage of the unique property of LiDAR point clouds. We adaptively assign hotspots according to different point-sparsity within each bounding box, which can be obtained from annotations. Whereas in RGB images CNNs tend to learn from texture information \cite{geirhos2018imagenettrained}, from which it is hard to measure how rich the features are in each object.
\subsubsection{Anchor-Free Detectors for Point Clouds} 
Some algorithms without anchors are proposed for indoors scenes. SGPN \cite{wang2018sgpn} segments instances by semantic segmentation and learning a similarity matrix to group points together. This method is not scalable since the size of similarity matrix grows quadratically with the number of points. 3D-BoNet \cite{yang2019learning} learns bounding boxes to provide a boundary for points from different instances. Unfortunately, both methods will fail when only partial point clouds have been observed, which is common in LiDAR point clouds. PIXOR \cite{yang2018pixor} and LaserNet \cite{meyer2019lasernet} project LiDAR points into bird's eye view (BEV) or range view and use standard 2D CNNs to produce bounding boxes in BEV. Note that we do not count VoteNet \cite{qi2019deep} and Point-RCNN \cite{shi2019pointrcnn} as anchor-free methods due to usage of anchor sizes.

\subsubsection{Efforts Addressing Regression Target Imbalance}\label{efforts}
The bounding box centers and sizes appear in different scales. Some objects have relatively large sizes while others do not. The scale variances in target values give rise to the scale variances in gradients. Small values tend to have smaller gradients and have less impact during training. Regression target imbalance is a great challenge for anchor-free detectors. Anchor-free detectors \cite{tian2019fcos,zhu2019feature,duan2019centernet,law2018cornernet,zhou2019objects,kong2019foveabox} became popular after Feature Pyramid Networks (FPN) \cite{lin2017feature} was proposed to handle objects of different sizes.

Complimentary to FPNs, anchor-based detectors \cite{ren2015faster,girshick2015fast,lin2017focal,liu2016ssd} rely on anchor locations and sizes to serve as normalization factors to guarantee that regression targets are mostly small values around zero. Multiple sizes and aspect ratios are hand-designed to capture the multi-modal distribution of bounding box sizes. Anchor-free detectors can be regarded as anchor-based detectors with one anchor of unit size at each location and thus anchor-free detectors don't enjoy the normalizing effect of different anchor sizes.
\section{Object as Hotspots} 
\subsection{Hotspot Definition}\label{hotspot_def}
We represent an object as composition of hotspots. \textbf{Spots} are defined as non-empty voxels which have points and overlap with objects. Only a subset of spots are assigned as \textbf{hotspots} and used for training, to mitigate the imbalance of number of points and the effect of missing or occluded part of objects. Hotspots are responsible for aggregating minimal and the most discriminative features of an object for background/foreground or inter-class classification. In training, hotspots are assigned by ground truth; in inference, hotspots are predicted by the network.

Intuitively the hotspots should satisfy three properties: 1) they should compose distinguishable parts of the objects in order to capture discriminative features; 2) they should be shared among objects of the same category so that common features can be learned from the same category; 3) they should be minimal so that when only a small number of LiDAR points are scanned in an object, hotspots still contain essential information to predict semantic information and localization, i.e. hotspots should be robust to objects with a small number of points.

\subsection{Hotspot Selection \& Assignment} \label{hotspot}

Hotspot selection \& assignment is illustrated in Fig. \ref{fig:reg_targets} (a). Unlike previous anchor-free detectors \cite{yang2018pixor,zhou2019objects}, which densely assign positive samples inside objects, we only select a subset of spots on objects as hotspots. We assign hotspots to the output feature map of the backbone network. After passing through the backbone network, a neuron on the feature map can be mapped to a super voxel in input point cloud space. We denote a voxel corresponding to a neuron on the output feature map as $V_n$, where $n$ indexes a neuron.

The annotations do not tell which parts are distinguishable, but we can infer them from the ground truth bounding boxes $B_{gt}$. We assume $V_n$ is an interior voxel of the object if inside $B_{gt}$. Then we consider $V_n$ as a spot if it's both non-empty and inside $B_{gt}$. We choose hotspots as nearest spots to the object center based on two motivations: 1) Points away from the object center are less reliable compared to those near the object centers, i.e., they are more vulnerable to the change of view angle.  2) As stated in FCOS~\cite{tian2019fcos}, locations closer to object centers tend to provide more accurate localization. 

We choose at most $M$ nearest spots as hotspots in each object. $M$ is an adaptive number determined by $M=\frac{C}{Vol}$, where $C$ is a hyperparameter we choose and $Vol$ is the volume of the bounding box. Because relatively large objects tend to have more points and richer features, we use $M$ to further suppress the number of hotspots in these objects. If the number of spots in an object is less than $M$, we assign all spots as hotspots.

\begin{figure*}[h]
\includegraphics[width=\linewidth]{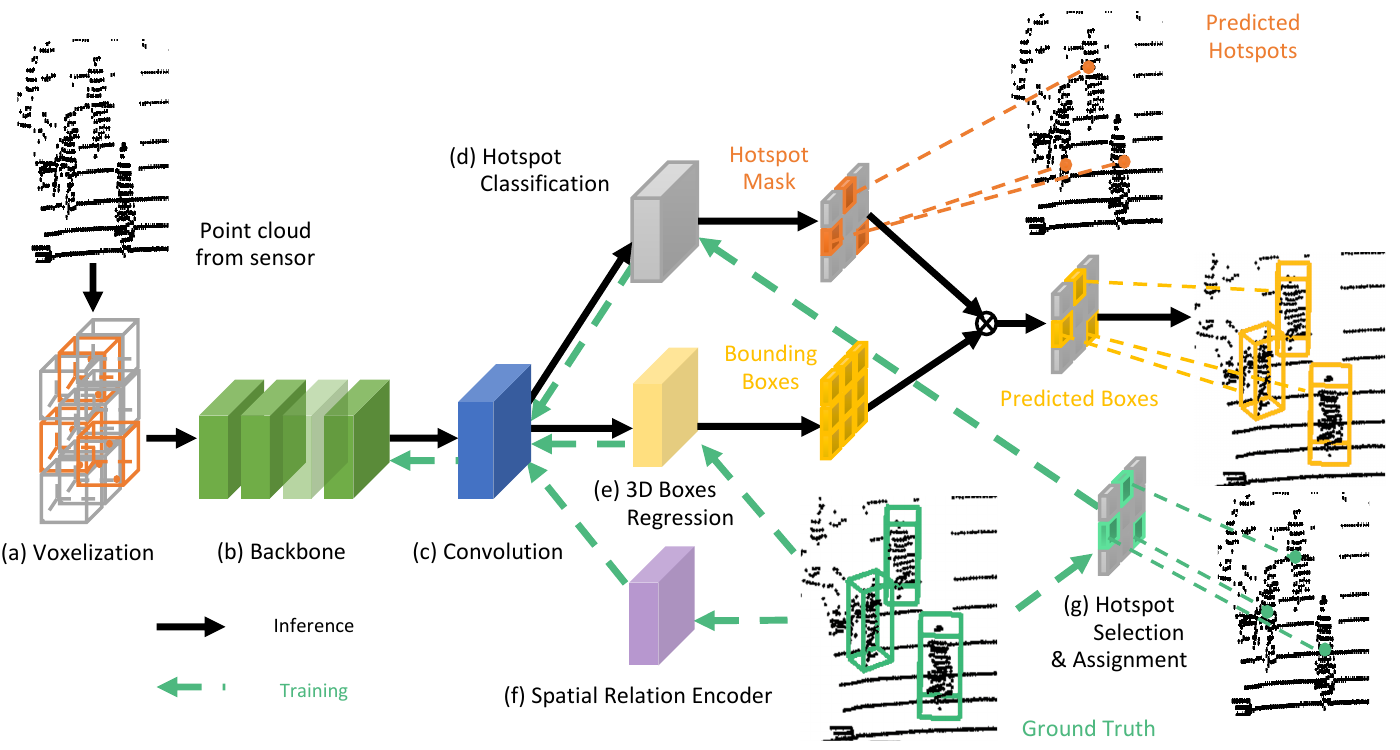}
\caption{Outline of HotSpotNet. The point cloud is (a) voxelized, and passed through the (b) backbone network to produce 3D feature maps.  These feature maps go through (c) a shared convolution layer, pass into three modules to perform (d) Hotspot Classification and (e) 3D Bounding Box regression (f) Spatial Relation Encoder to train the network, and (g) selected hotspots are assigned as positive labels to (d) Hotspot Classification. During inference only (d) Hotspot Classification and (e) 3D Bounding Box Regression are performed to obtain hotspots and bounding boxes respectively.
}
\label{fig:subnets}
\end{figure*}
\section{HotSpot Network} \label{network}

Based on OHS, we architect the Hotspot Network (HotSpotNet) for LiDAR point clouds. HotSpotNet consists of a 3D feature extractor and Object-as-Hotspots (OHS) head. OHS head has three subnets for hotspot classification, box regression and spatial relation encoder. 

The overall architecture of our proposed HotSpotNet is shown in Fig. \ref{fig:subnets}. The input LiDAR point clouds are voxelized into cuboid-shape voxels. The input voxels pass through the 3D CNN to generate the feature maps. The three subnets will guide the supervision and generate the predicted 3D bounding boxes. Hotspot assignment happens at the last convolutional feature maps of the backbone. The details of network architecture and the three subnets for supervision are described below. 

\subsection{Object-as-Hotspots Head}
Our OHS head network consists of three subnets: 1) a hotspot classification subnet that predicts the likelihood of class categories; 2) a box regression subnet that regresses the center locations, dimensions and orientations of the 3D boxes. 3) a spatial relation encoder for hotspots.

\subsubsection{Hotspot Classification} The classification module is a convolutional layer with $K$ heatmaps each corresponding to one category. The hotspots are labeled as ones. The targets for all the non-hotspots are zeros. We apply a gradient mask so that gradients for \textit{non-hotspots inside the ground truth bounding boxes} are set to zero. That means they are ignored during training and do not contribute to back-propagation. Binary classification is applied to hotspots and non-hotspots. Focal loss \cite{lin2017focal} is applied at the end, 

\begin{equation} \label{loss}
\begin{split}
&\mathcal{L}_{cls} = \sum_{k=1}^{K} \alpha (1-p_{k})^{\gamma}log(p_{k})\\
\end{split}
\end{equation}
where, $$ p_{k}=\left\{
\begin{array}{rcl}
 &p            &  ,\text{hotspots} \\
 &(1 - p)    &, \text{non-hotspots} \\
\end{array} \right. $$

$p$ is the output probability, and $K$ is the number of categories. The total classification loss is averaged over the total number of hotspots and non-hotspots, excluding the non-hotspots within ground truth bounding boxes.

\begin{figure*}[h]
\begin{subfigure}
  \centering
\includegraphics[width=0.4\linewidth]{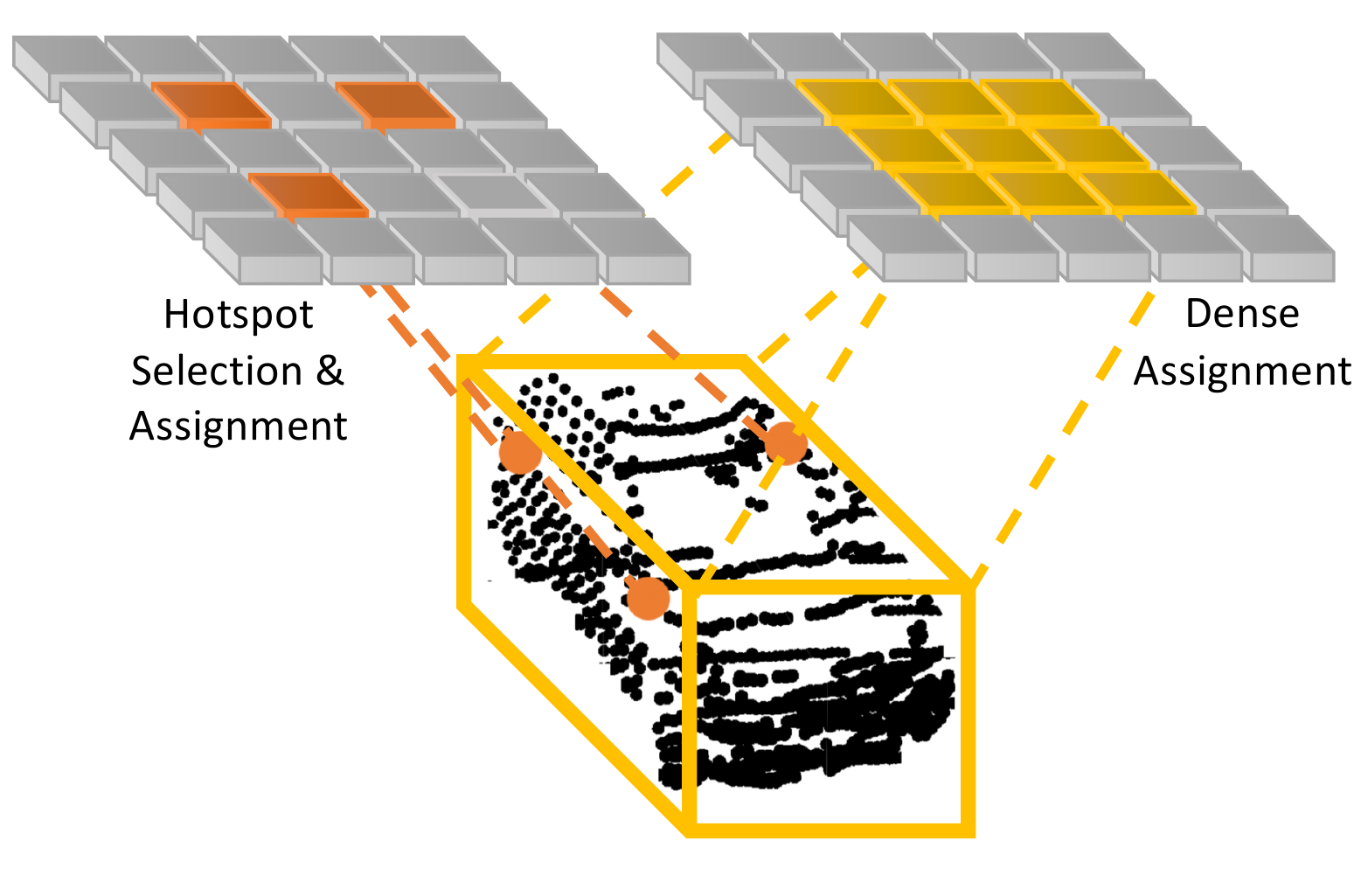}
\end{subfigure}%
\begin{subfigure}
 \centering
 \includegraphics[width=0.27\linewidth]{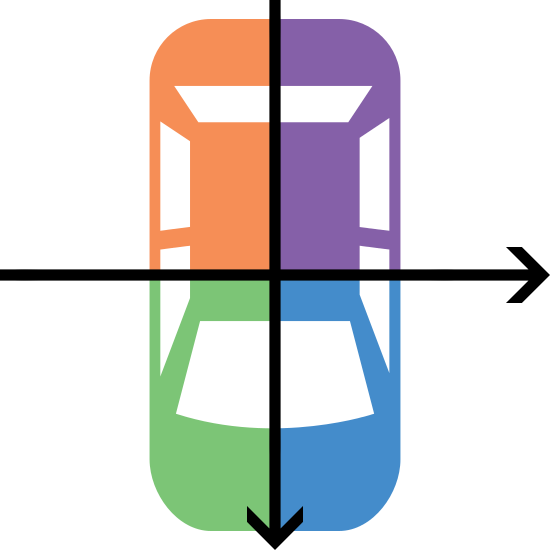}
\end{subfigure}%
\begin{subfigure}
 \centering
 \includegraphics[width=0.27\linewidth]{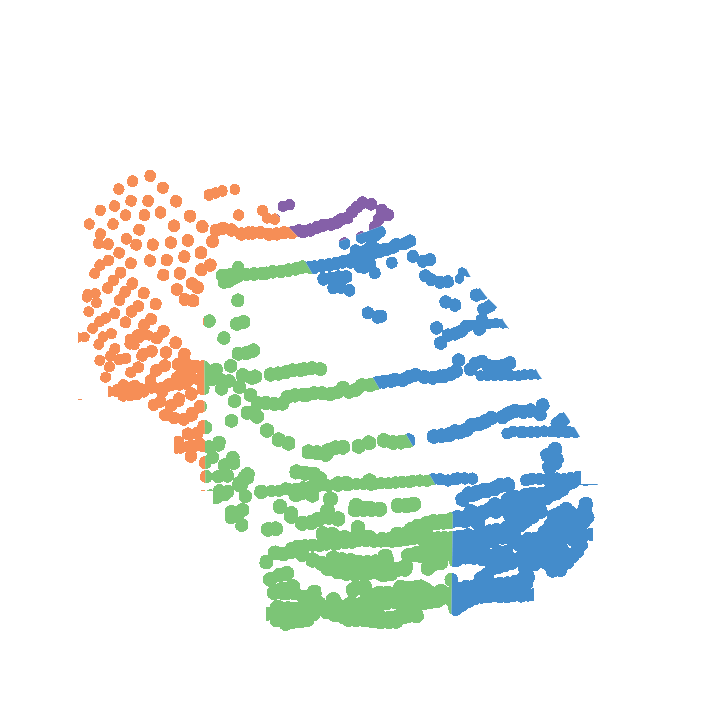}
\end{subfigure}

\caption{Left: illustration of hotspot selection \& assignment. Only selected non-empty voxels on objects are assigned as hotspots. Previous anchor-free detectors \cite{tian2019fcos,zhu2019feature} densely assign locations inside objects as positive samples. Middle: spatial relation encoding: we divide the object bounding box in BEV into quadrants by the orientation (front-facing direction) and its perpendicular direction. Quadrants I, II, III, and IV are color-coded with green, blue, purple and orange respectively in the illustration. Right: illustration of how points of a vehicle are classified into different quadrants, with the same set of color-coding as the middle figure.}
\label{fig:reg_targets}
\end{figure*}

\subsubsection{Box Regression} The bounding box regression only happens on hotspots. For each hotspot, an eight-dimensional vector $[d_x, d_y, z, \log(l), \log(w), \log(h), \cos(r),\\ \sin(r)]$ is regressed to represent the object in LiDAR point clouds. $d_x, d_y$ are the axis-aligned deviations from the hotspot to the object centroid. The hotspot centroid in BEV can be obtained by:
\begin{equation} \label{hot_spot_location}
\begin{split}
[x_h, y_h]=(\frac{j+0.5}{L}(x_{max}-x_{min})+x_{min}, \\ \frac{i+0.5}{W}(y_{max}-y_{min})+y_{min}), 
\end{split}
\end{equation}
where $i, j$ is the spatial index of its corresponding neuron on the feature map with size ${W \times L}$, and $[x_{min}, x_{max}]$, $[y_{min}, y_{max}]$ are the ranges for $x, y$ when we voxelize all the points. 

As discussed in Sec. \ref{efforts}, anchor-free detectors suffer from regression target imbalance. Instead of introducing FPN, i.e. extra layers and computational overhead to our network, we tackle regression target imbalance by carefully designing the targets: 1) We regress $\log(l)$, $\log(w)$, $\log(h)$ instead of their original values because $\log$ scales down the absolute values; 2) We regress $\cos(r)$, $\sin(r)$ instead of $r$ directly because they are constrained in $[-1, 1]$ instead of the original angle value in $[-\pi, \pi]$; 3) We use soft $argmin$ \cite{kendall2017end} to help regress $d_x$, $d_y$ and $z$. To regress a point location in a segment ranging from $a$ to $b$ by soft $argmin$, we divide the segment into $N$ bins, each bin accounting for a length of $\frac{b-a}{N}$. The target location can be represented as $t = \Sigma_i^N (S_i C_i)$, where $S_i$ represents the softmax score of the $i_{th}$ bin and $C_i$ is the center location of the $i_{th}$ bin. Soft $argmin$ is widely used in stereo vision to predict disparity in sub-pixel resolution. We notice soft $argmin$ can address regression target imbalance by turning the regression into classification problem and avoiding regressing absolute values.

Smooth L1 loss \cite{girshick2015fast} is adopted for regressing these bounding box targets and the regression loss is only computed over hotspots.
\begin{equation} \label{loss}
\mathcal{L}_{loc} (x) =\left\{
\begin{array}{rcl}
&0.5x^2       &  ,|x| < 1 \\
&|x| - 0.5    &, \text{otherwise} \\
\end{array} \right. 
\end{equation}
\subsubsection{Spatial Relation Encoder} 

Inspired by compositional models, we incorporate hotspot spatial relations to our HotSpotNet. Since convolution is translation-invariant, it's hard for a CNN to learn spatial relations without any supervision. Therefore, we explore the implicit spatial relation from annotations. We observe that most target objects for autonomous driving can be considered as rigid objects (e.g. cars), so the relative locations of hotspots to object centers do not change, which can be determined with the help of bounding box centers and orientations. We thus categorize the relative hotspot location to the object center on BEV into a one-hot vector representing quadrants, as shown in Fig. \ref{fig:reg_targets} Middle\&Right. We train hotspot spatial relation encoder as quadrant classification with binary cross-entropy loss and we compute the loss only for hotspots.

\begin{equation} \label{loss}
\mathcal{L}_{q} = \sum^{3}_{i=0}-[q_i\log(p_i) + (1-q_i)\log(1-p_i)]
\end{equation}
where $i$ indexes the quadrant, $q_i$ is the target and $p_i$ the predicted likelihood falling into the specific quadrant. 

\subsection{Learning and Inference} 

The final loss for our proposed HotSpotNet is the weighted sum of losses from three branches:
\begin{equation} \label{loss}
\begin{split}
\mathcal{L} = \delta \mathcal{L}_{cls} + \beta \mathcal{L}_{loc} + \zeta \mathcal{L}_q
\end{split}
\end{equation}

Where, $\delta$, $\beta$ and $\zeta$ are the weights to balance the classification, box regression and spatial relation encoder loss. 

During inference, if the corresponding largest entry value of the $K$-dimensional vector of the classification heatmaps is above the threshold, we consider the location as hotspot firing for the corresponding object. Since one instance might have multiple hotspots, we further use Non-Maximum Supression (NMS) with the Intersection Over Union (IOU) threshold to pick the most confident hotspot for each object. The spatial relation encoder does not contribute to inference.

\section{Experiments}
\label{expe}
In this section, we summarize the dataset in Sec. \ref{dataset} and present the implementation details of our proposed HotSpotNet in \ref{details}. We evaluate our method on KITTI 3D detection Benchmark \cite{geiger2012we} in Sec. \ref{benchmark} and NuScenes 3D detection dataset \cite{caesar2019nuscenes} in Sec. \ref{nusc}. We also analyze the advantages of HotSpotNet in Sec. \ref{compare_distance} and present ablation studies in Sec. \ref{ablation}. 

\subsection{Datasets and Evaluation}\label{dataset}
\textbf{KITTI Dataset} \quad KITTI has 7,481 annotated LiDAR point clouds for training with 3D bounding boxes for classes such as cars, pedestrians and cyclists. It also provides 7,518 LiDAR point clouds for testing. In the rest of paper, all the ablation studies are conducted on the common train/val split, i.e. 3712 LiDAR point clouds for training and 3769 LiDAR point clouds for validation. To further compare the results with other approaches on KITTI 3D detection benchmark, we randomly split the KITTI annotated data into $4:1$ for training and validation and report the performance on KITTI test dataset. Following the official KITTI evaluation protocol, average precision (AP) based on 40 points is applied for evaluation. The IoU threshold is 0.7 for cars and 0.5 for pedestrians and cyclists.

\textbf{NuScenes Dataset} \quad The dataset contains $1,000$ scenes, including $700$ scenes for training, $150$ scenes for validation and $150$ scenes for test. $40, 000$ frames are annotated in total, including 10 object categories. The mean average precision (mAP) is calculated based on the distance threshold (i.e. $0.5m, 1.0m, 2.0m$ and $4.0m$). Additionally, a new metric, nuScenes detection score (NDS) \cite{caesar2019nuscenes}, is introduced as a weighted sum of mAP and precision on box location, scale, orientation, velocity and attributes. 

\subsection{Implementation Details}\label{details}
\textbf{Backbone Network}
In experiments on KITTI, we adopt the same backbone as used by SECOND \cite{yan2018second}. We set point cloud range as $[0,70,4]$, $[-40,40]$, $[-3,1]$ and voxel size as $(0.025, 0.025, 0.05)m$ along $x,y,z$ axis. A maximum of five points are randomly sampled from each voxel and voxel features are obtained by averaging corresponding point features. 

As for NuScenes, we choose the state-of-the-art method CBGS \cite{zhu2019class} as our baseline. Input point cloud range is set to $[-50.4,50.4], [-50.4,50.4], [-5,3]$ along $x, y, z$, respectively. We implement our method with ResNet \cite{he2016deep} and PointPillars (PP) \cite{lang2019pointpillars} backbones and report each performance. We set voxel size as $(0.1,0.1,0.16)m$ for ResNet backbone and $(0.2,0.2)m$ for PP backbone. For each hotspot, we also set $(\log l, \log w, \log h)$ as outputs of soft $argmin$ to handle the size variances for 10 object categories.

\textbf{Object-as-Hotspots Head}
Since the output feature map of the backbone network is consolidated to BEV, in this paper we assign hotspots in BEV as well. Our OHS head consists of a shared $3\times 3$ convolution layer with stride $1$. We use a $1\times 1$ convolution layer followed by sigmoid to predict confidence for hotspots. For regression, we apply several $1\times 1$ convolution layers to different regressed values. Two $1\times 1$ convolution layers are stacked to predict soft $argmin$ for $(d_x$, $d_y)$ and $z$. Additional two $1\times 1$ convolution layers to predict the dimensions and rotation. We set the range $[-4, 4]$ with 16 bins for $d_x$, $d_y$ and 16 bins for $z$, with the same vertical range as the input point cloud. We set $C=64$ to assign hotspots. For hotspot spatial relation encoder, we use another a $1\times 1$ convolution layer with softmax for cross-entropy classification. 
We set $\gamma=2.0$ and $\alpha=0.25$ for focal loss. For KITTI,  the loss weights are set as $\delta=\beta=\zeta=1$. For NuScenes we set $\delta=1$ and $\beta=\zeta=0.25$.

\textbf{Training and Inference} For KITTI, we train the entire network end-to-end with adamW \cite{loshchilov2017fixing} optimizer and one-cycle policy \cite{smith2019super} with LR max $2.25e^{-3}$, division factor $10$, momentum ranges from $0.95$ to $0.85$ and weight decay $0.01$. We train the network with batch size 8 for 150 epochs. During testing, we keep 100 proposals after filtering the confidence lower than $0.3$, and then apply the rotated NMS with IOU threshold $0.01$ to remove redundant boxes. 

For NuScenes, we set LR max as $0.001$. We train the network with batch size 48 for 20 epochs. During testing, we keep 80 proposals after filtering the confidence lower than $0.1$, and IOU threshold for rotated NMS is $0.02$.

\textbf{Data Augmentation} Following SECOND\cite{yan2018second}, for KITTI, we apply random flipping, global scaling, global rotation, rotation and translation on individual objects, and GT database sampling. For NuScenes, we adopt same augmentation strategies as in CBGS \cite{zhu2019class} except we add random flipping along $x$ axis and attach GT objects from the annotated frames. Half of points from GT database are randomly dropped and GT boxes containing fewer than five points are abandoned. 

\subsection{Experiment results on KITTI benchmark}\label{benchmark}
As shown in Table \ref{kitti-test}, we evaluate our method on the KITTI test dataset. For fair comparison, we also show the performance of our implemented SECOND \cite{yan2018second} with same voxel size as ours, represented by HR-SECOND in the table. For the 3D object detection benchmark, solely LiDAR-based, our proposed HotSpotNet outperforms all published LiDAR-based, one-stage detectors on cars, cyclists and pedestrians of all difficulty levels. In particular, by the time of submission our method ranks 1$st$ among all published methods on KITTI test set for cyclist and pedestrian detection. HotSpotNet shows its advantages on objects with a small number of points. The results demonstrate the success of representing objects as hotspots. Our one-stage approach also beats some classic 3D two-stage detectors for car detection, including those fusing LiDAR and RGB images information. Still, our proposed OHS detection head is complimentary to architecture design in terms of better feature extractors.  
\subsubsection{Inference Speed} The inference speed of HotSpotNet is 25FPS, tested on KITTI dataset with a Titan V100. We compare inference speed with other approaches in Table \ref{kitti-test}. We achieve significant performance gain while maintaining the speed as our baseline SECOND \cite{yan2018second}.

\begingroup
\renewcommand{\arraystretch}{0.8}
\begin{table*}[ht]
\begin{center}
\scalebox{0.95}[1.0]{
\begin{tabular}{c|c|c|c|ccc|ccc|ccc}
\hline
\multirow{2}{*}{Method} & \multirow{2}{*}{Input} & \multirow{2}{*}{Stage} &
\multirow{2}{*}{FPS} &
\multicolumn{3}{c|}{3D Detection (Car)} & \multicolumn{3}{c|}{3D Detection (Cyclist)} & \multicolumn{3}{c}{3D Detection (Pedestrian)}  \\
 &  &  & & Mod & Easy & Hard & Mod & Easy & Hard & Mod & Easy & Hard \\ \hline
ComplexYOLO\cite{simon2018complex} & L & One &17 & 47.34 & 55.93 & 42.60 & 18.53 & 24.27 & 17.31 & 13.96 &	17.60 &	12.70 \\
VoxelNet\cite{zhou2018voxelnet} & L & One &4 & 65.11 & 77.47 & 57.73 & 48.36 & 61.22 & 44.37 & 39.48 & 33.69 & 31.51 \\
SECOND-V1.5\cite{yan2018second} & L & One &33 & 75.96 & 84.65 & 68.71 & - & - & - & - & - & - \\
HR-SECOND\cite{yan2018second} & L & One &25 & 75.32 & 84.78 & 68.70 & 60.82 & 75.83 & 53.67 & 35.52 & 45.31 & 33.14 \\
PointPillars\cite{lang2019pointpillars} & L & One &62 & 74.31 & 82.58 & 68.99  & 58.65 & 77.10 & 51.92 & 41.92 & 51.45 & 38.89 \\
3D IoU Loss\cite{zhou2019iou} & L & One &13 & 76.50 & 86.16 & 71.39 & - & - & - & - & - & - \\
HRI-VoxelFPN\cite{wang2019voxel} & L & One &50 & 76.70 & 85.64 & 69.44 & - & - & - & - & - & - \\
ContFuse \cite{liang2018deep} & I + L & One &17 & 68.78 & 83.68 & 61.67 &  - & - & - & - & - & - \\ \hline
MV3D \cite{chen2017multi} & I + L & Two &3 &63.63 & 74.97 & 54.00 & - & - & - & - & - & - \\
AVOD-FPN \cite{ku2018joint} & I + L & Two &10 & 71.76 & 83.07 & 65.73 & 50.55 & 63.76 & 44.93 & 42.27 & 50.46 & 39.04 \\
F-PointNet \cite{qi2018frustum} & I + L & Two &6 & 69.79 & 82.19 & 60.59 & 56.12 & 72.27 & 49.01 & 42.15 & 50.53 & 38.08 \\
F-ConvNet \cite{wang2019frustum} & I + L & Two &2 &76.39 & 87.36 & 66.69 &65.07 &81.89 &56.64 &43.38 &52.16 &38.8\\
MMF \cite{liang2019multi} & I + L & Two &13 & 77.43  & \color{blue}{88.40} & 70.22 & - & - & - & - & - & - \\
PointRCNN \cite{shi2019pointrcnn} & L & Two &10 & 75.64 & 86.96 & 70.70 & 58.82 & 74.96 & 52.53 & 39.37 & 47.98 & 36.01 \\ 
FastPointRCNN\cite{Chen_2019_ICCV}  & L & Two &17 &77.40	&85.29	&70.24	& - & - & - & -  & - & - \\
STD \cite{yang2019std} & L & Two  &13 & \color{blue}{79.71} & 87.95 & \color{blue}{75.09} & 61.59 & 78.69 & 55.30 &42.47 &53.29 &38.35\\ \hline
HotSpotNet & L & One & 25 & \textbf{78.31} & \textbf{87.60}  & \textbf{73.34} & \textbf{65.95} & \textbf{82.59} & \textbf{59.00} & \textbf{45.37} & \textbf{53.10}  & \textbf{41.47 }\\ 
\hline
\end{tabular}
}
\caption{
Performance of 3D object detection on KITTI test set. ``L", ``I" and ``L+I" indicates the method uses LiDAR point clouds, RGB images and fusion of two modalities, respectively. FPS stands for frame per second. Bold numbers denotes the best results for single-modal one-stage detectors. Blue numbers are results for best-performing detectors. 
} \label{kitti-test}
\end{center}
\end{table*}
\endgroup

\begingroup
\renewcommand{\arraystretch}{0.8}
\begin{table*}[]
\begin{center}
\scalebox{1.0}[1.0]{
\begin{tabular}{c|cccccccccc|c|c}
\hline
Method  & \footnotesize{car} & \footnotesize{truck} & \footnotesize{bus} & \footnotesize{trailer} & \shortstack[c]{ \scriptsize{constr-}\\\scriptsize{uction}\\\scriptsize{vehicle}} & \shortstack[c]{\vspace{0.5mm}\scriptsize{pede-}\\\scriptsize{strian}}  & \shortstack[c]{\vspace{0.5mm}\scriptsize{motor-} \\ \scriptsize{cycle}} \ & \footnotesize{bike} & \shortstack[c]{\vspace{0.5mm}\scriptsize{traffic} \\ \scriptsize{cone}} & \shortstack[c]{\vspace{0.5mm}\scriptsize{barr-} \\ \scriptsize{ier}} & \footnotesize{mAP} & \footnotesize{NDS} \\ \hline
CBGS-PP &81.3	&49.7&59.0	&32.1	&13.4	&73.1 &51.5	&23.5	&\textbf{51.3}	&\textbf{52.6} &48.8 &59.2	\\
HotSpotNet-PP &\textbf{83.3}	&\textbf{52.7}	&\textbf{63.7}	&\textbf{35.3}	&\textbf{15.3}	&\textbf{74.8}	&\textbf{53.7}	&\textbf{25.5}	&50.3	&52.0	&\textbf{50.6} &\textbf{59.8} \\
\hline
CBGS-ResNet &	82.9&	52.9	&64.6	&37.5	&18.3	&80.3	&60.1	&39.4	&64.8	&61.8 &56.3 &62.8	\\
HotSpotNet-ResNet &\textbf{84.0}	&\textbf{56.2}	&\textbf{67.4}	&\textbf{38.0}	&\textbf{20.7}	&\textbf{82.6}	&\textbf{66.2}	&\textbf{49.7}	&\textbf{65.8}	&\textbf{64.3} &\textbf{59.5} &\textbf{66.0}\\
\hline
\end{tabular}
}
\caption{
3D object detection mAP on NuScenes val set.
}\label{nusc-val}
\end{center}
\end{table*}
\endgroup

\subsection{Experiment results on NuScenes dataset}\label{nusc}

We present results on NuScenes validation set (Table \ref{nusc-val}) and test set (Table \ref{nusc-test}). We reproduced the baseline CBGS \cite{zhu2019class} based on implementation from CenterPoint \cite{yin2020center} without double-flip testing. Our reproduced mAPs are much higher than the results presented in the original CBGS paper. As shown in Table \ref{nusc-val}, our HotSpotNet outperforms CBGS by $1.8$ and $3.2$ in mAP for the PointPillars and ResNet backbone respectively. In Table \ref{nusc-test}, our approach outperforms all detectors on the NuScenes 3D Detection benchmark using a single model.
\begin{table*}[h]
  \label{test}
  \centering
  \scalebox{1.0}[1.0]{
  \begin{tabular}{c|cccccccccc|c|c}
  \hline
  Method  & \footnotesize{car} & \footnotesize{truck} & \footnotesize{bus} & \footnotesize{trailer} & \shortstack[c]{ \scriptsize{constr-}\\\scriptsize{uction}\\\scriptsize{vehicle}} & \shortstack[c]{\vspace{0.5mm}\scriptsize{pede-}\\\scriptsize{strian}}  & \shortstack[c]{\vspace{0.5mm}\scriptsize{motor-} \\ \scriptsize{cycle}} \ & \footnotesize{bike} & \shortstack[c]{\vspace{0.5mm}\scriptsize{traffic} \\ \scriptsize{cone}} & \shortstack[c]{\vspace{0.5mm}\scriptsize{barr-} \\ \scriptsize{ier}} & \footnotesize{mAP} & \footnotesize{NDS}\\
    \hline
    SARPNET \cite{ye2020sarpnet} &$59.9$ &$18.7$ &$19.4$ & $18.0$ &$11.6$ &$69.4$ &$29.8$ &$14.2$ &$44.6$ &$38.3$ &$31.6$  &$49.7$\\
    PointPillars \cite{lang2019pointpillars} &$68.4$ &$23.0$ &$28.2$ &$23.4$ &$4.1$ &$59.7$ &$27.4$ &$1.1$ &$30.8$ &$38.9$&$30.5$  &$45.3$\\
    WYSIWYG \cite{hu2019you} &$79.1$ &$30.4$ &$46.6$ &$40.1$ &$7.1$ &$65.0$ &$18.2$ &$0.1$ &$28.8$ &$34.7$&$35.0$  &$41.9$\\
    CBGS \cite{zhu2019class} &$81.1$ &$48.5$ &$54.9$ &$42.9$ &$10.5$ &$80.1$ &$51.5$ &$22.3$ &$70.9$ &$65.7$&$52.8$  &$63.3$\\
    HotSpotNet-ResNet (Ours) &$\textbf{83.1}$&$\textbf{50.9}$ &$\textbf{56.4}$ &$\textbf{53.3}$ & $\textbf{23.0}$&$\textbf{81.3}$&$\textbf{63.5}$ &$\textbf{36.6}$ &$\textbf{73.0}$ &$\textbf{71.6}$ &$\textbf{59.3}$  &$\textbf{66.0}$\\
    \hline
  \end{tabular}
  }\caption{3D detection mAP on the NuScenes test set}\label{nusc-test}
\end{table*}
\subsection{Analysis}
\begin{figure*}[h!]
  \centering
  \begin{subfigure}
 \centering
 \includegraphics[width=.33\linewidth]{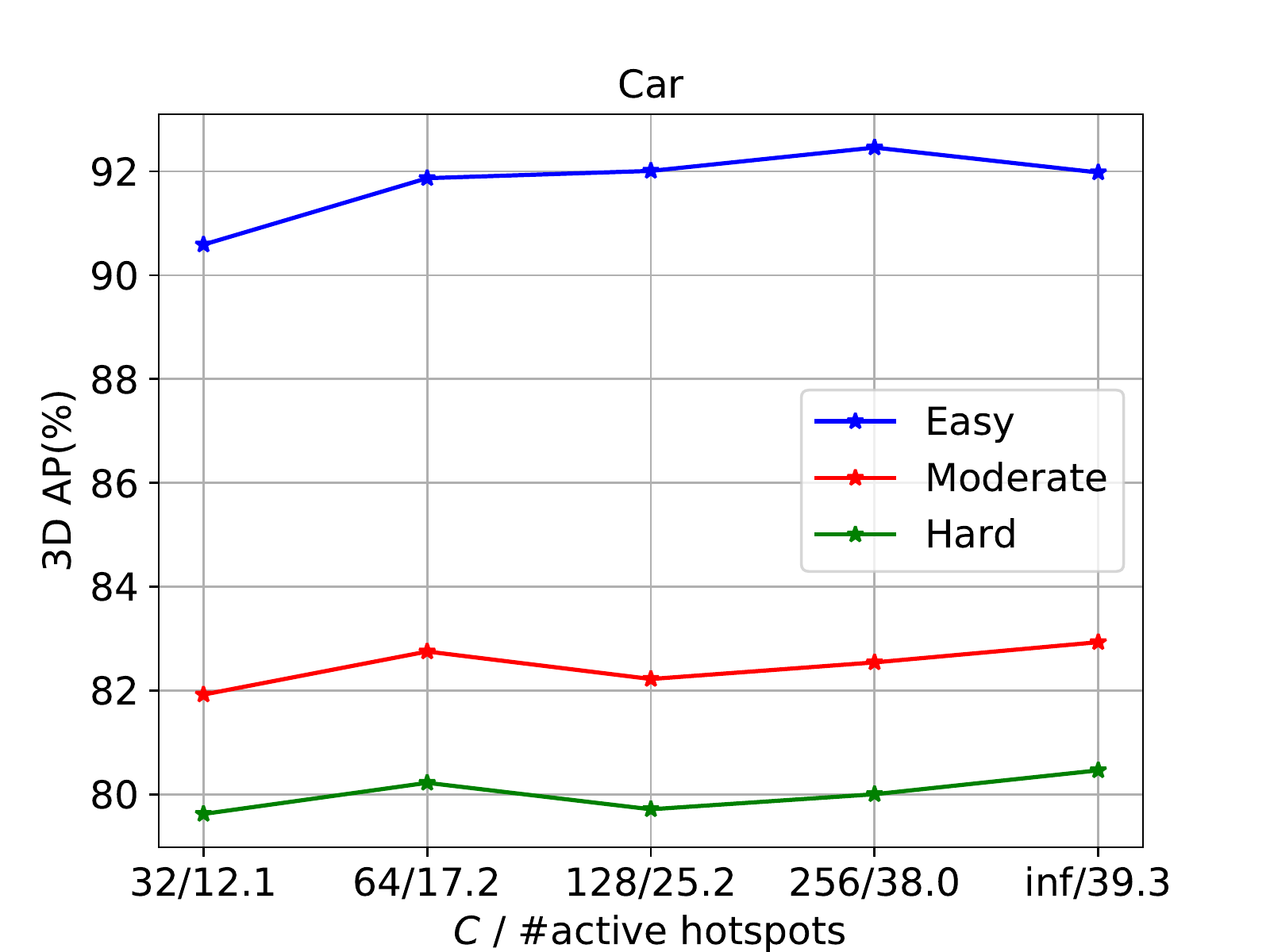}
 \label{fig:sub1}
\end{subfigure}%
\begin{subfigure}
 \centering
 \includegraphics[width=.33\linewidth]{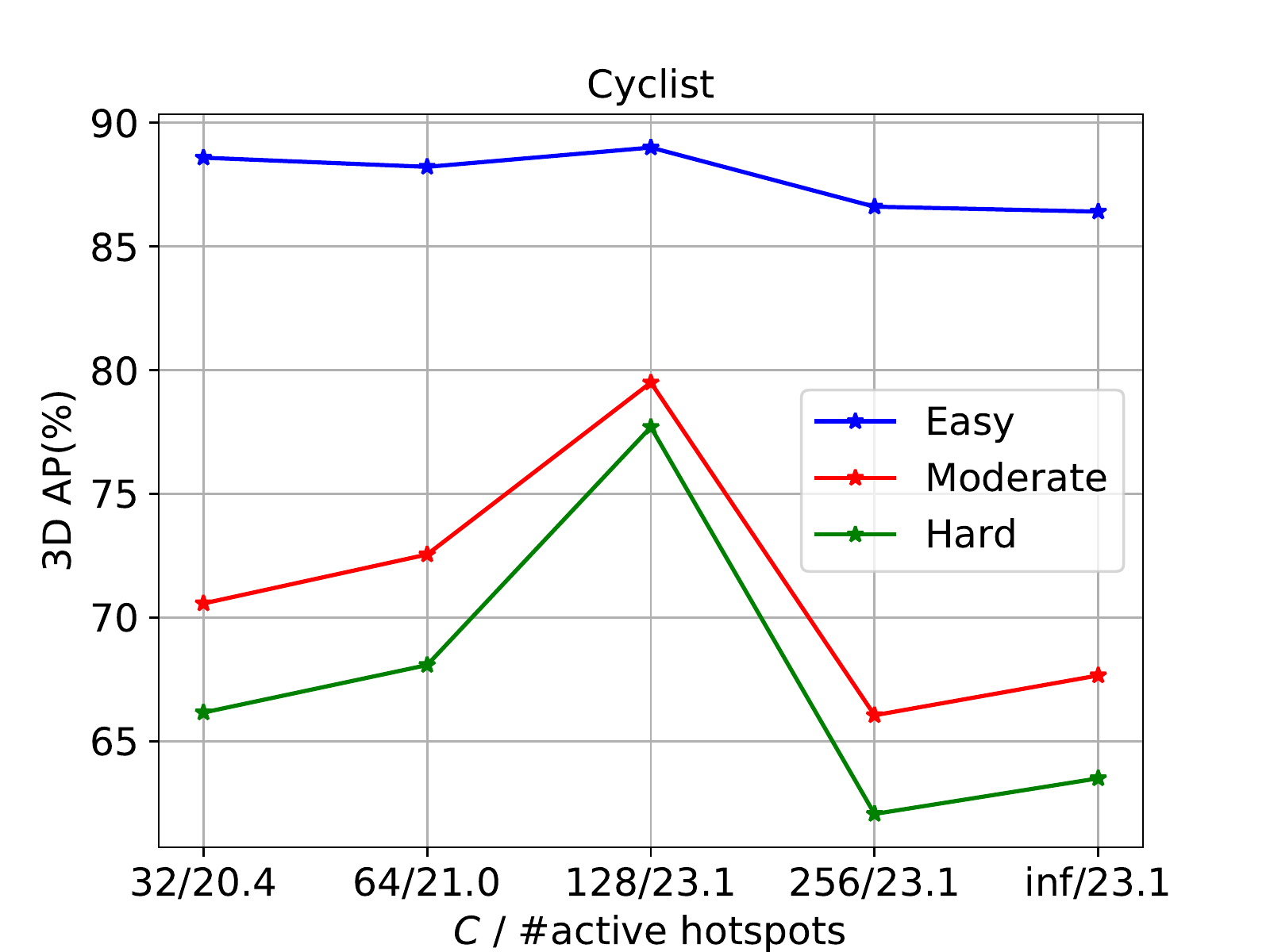}
 \label{fig:sub1}
\end{subfigure}%
  \begin{subfigure}
 \centering
 \includegraphics[width=.33\linewidth]{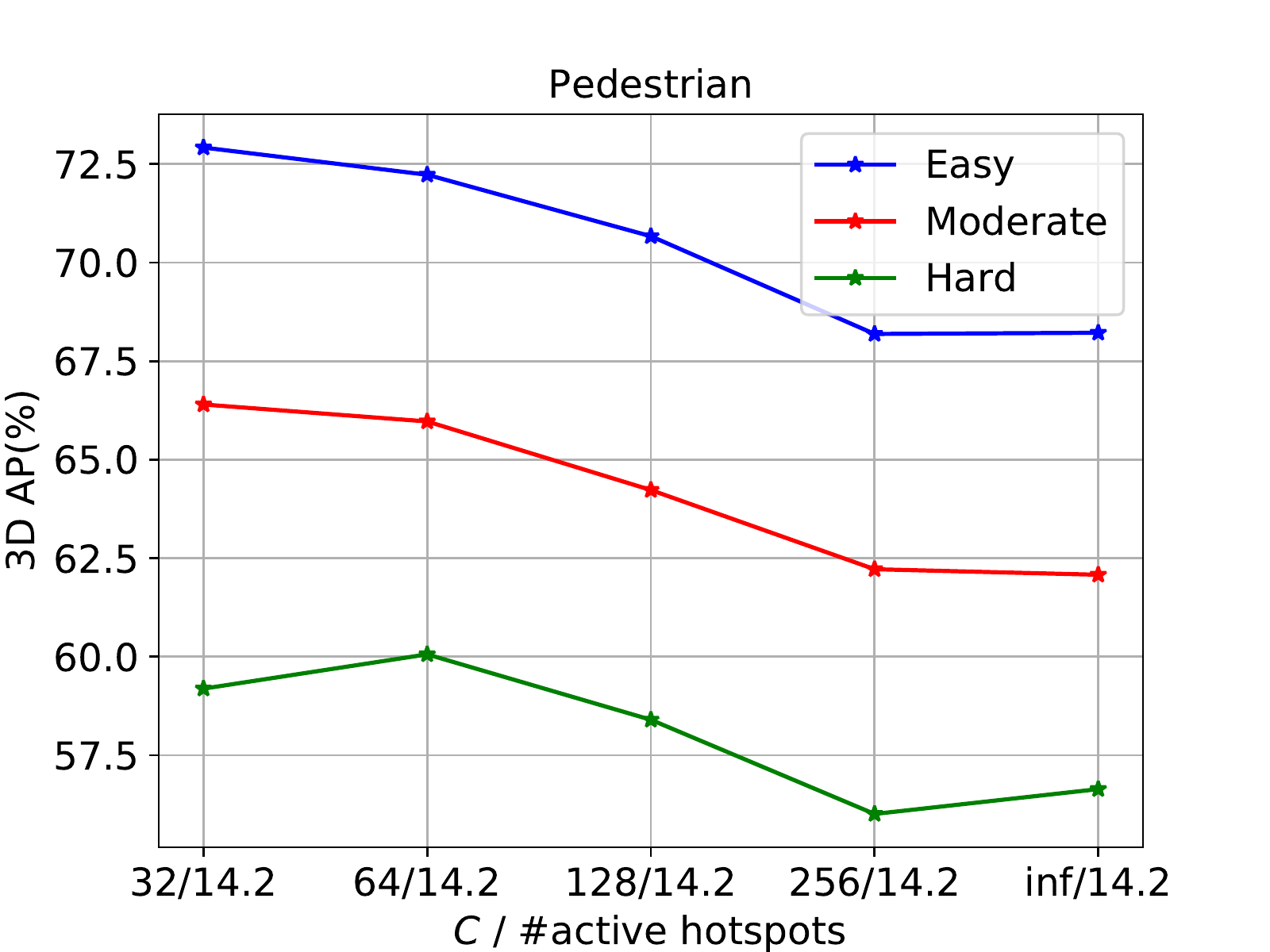}
 \label{fig:sub1}
\end{subfigure}%
\caption {Performances with different $C$ values on KITTI val. The horizontal axis also shows the number of active hotspots on average with different $C$ values.}\label{C}
\end{figure*} 

We argue that our approach advances in preventing the network from biasing towards objects with more points without compromising performance on these objects. We analyze the effect of different number of hotspots and performance on objects with different number of points.

\subsubsection{Different Number of Hotspots} 

In Sec. \ref{hotspot}, we set $M=\frac{C}{Vol}$ as the maximum number of hotspots in each object during training. Here we present the performances with different $C$ values: $32, 64, 128, 256, Inf$, where $Inf$ means we assign all spots as hotspots. The results are shown in Fig. \ref{C}. We can see that generally the larger $C$ is, the higher performance in detecting cars. We only perceive a significant drop when $C=32$ and the overall performance in detecting cars is not sensitive to different values of $C$. The performance in detecting cyclists reaches its peak when $C=128$. The lower the $C$ value, the better performance in detecting pedestrians. The performance of detecting pedestrians does not change much when $C\leq 64$. To balance the performance on all classes and prevent over-fitting on one class, we choose $C=64$ in our paper.

\subsubsection{Performance on objects with different number of points}\label{compare_distance}
\begin{figure*}
\centering
  \begin{subfigure}
  \centering
\includegraphics[width=0.4\linewidth]{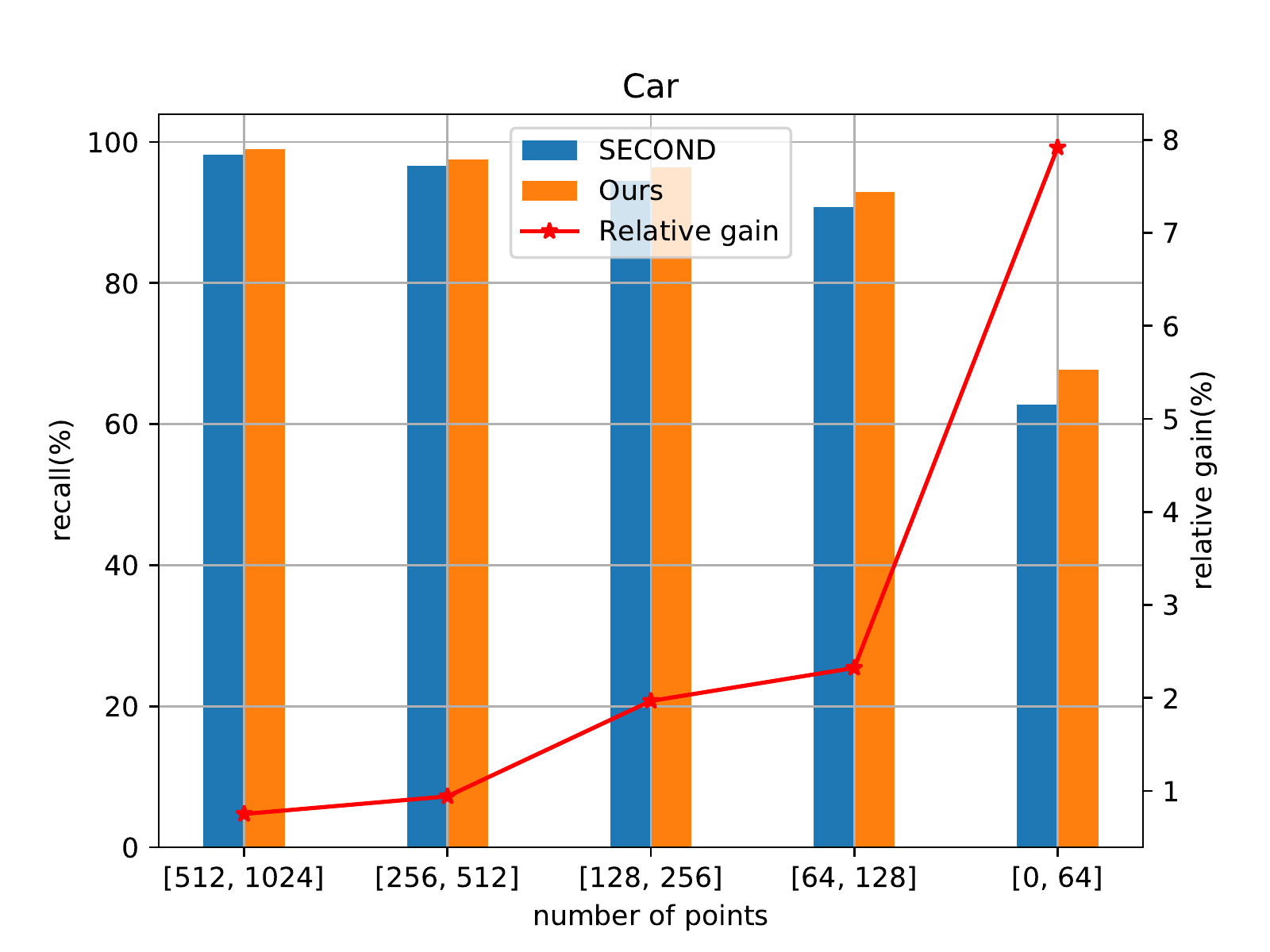}
\end{subfigure}%
\begin{subfigure}
  \centering
\includegraphics[width=0.4\linewidth]{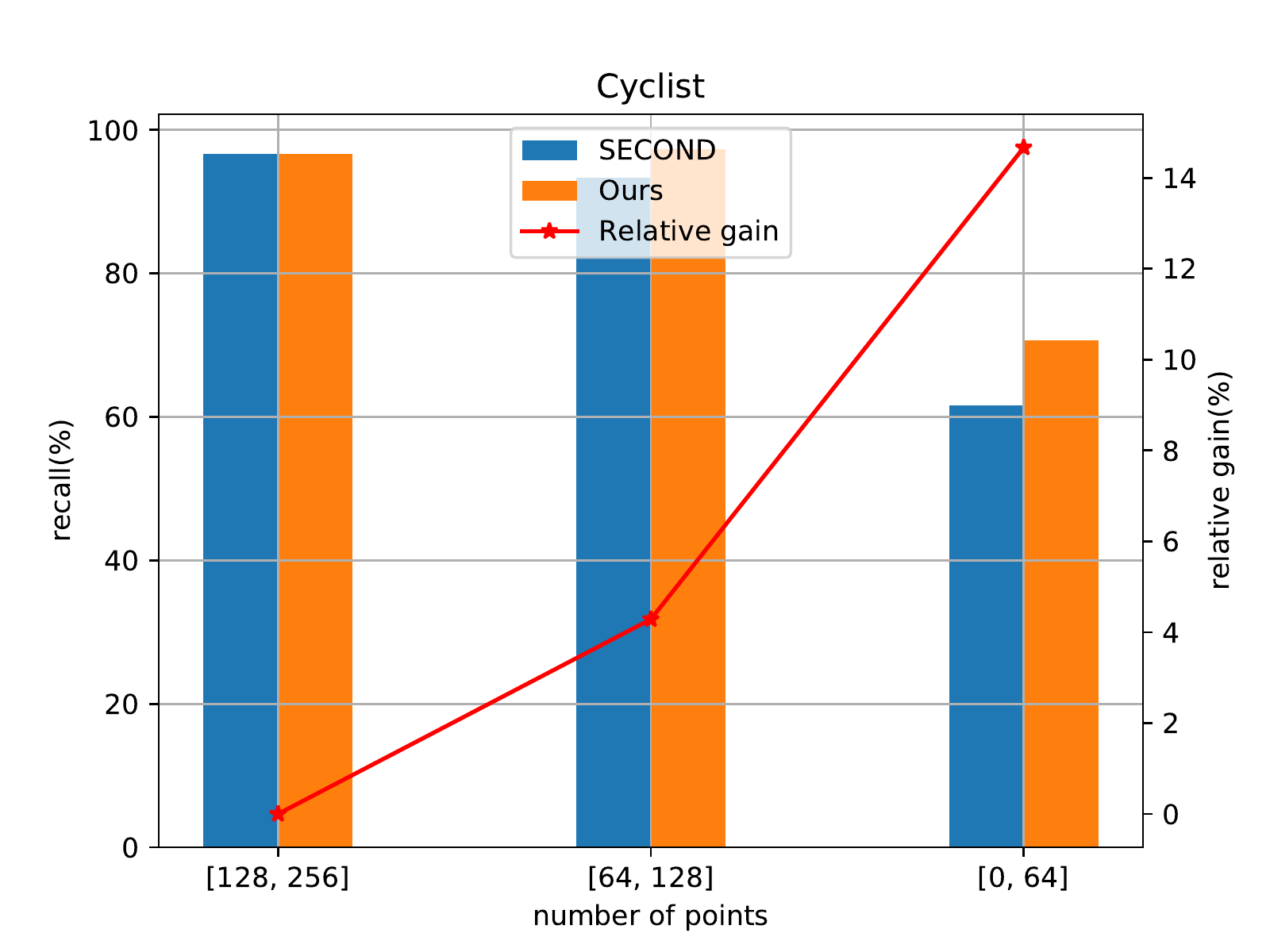}
\end{subfigure}%
 
\caption{Recall of detecting objects with different number of points on KITTI val.}
\label{distance}
\end{figure*}

Comparison between SECOND \cite{yan2018second} and our approach for objects with different number of points is shown in Fig. \ref{distance}. Our approach is consistently better to detect objects with different number of points and less likely to miss objects even with a small number of points. Notably, the relative gain of our approach compared to SECOND increases as the number of points decreases, showing our approach is more robust to sparse objects. 
\begin{table*}[h]
\begin{center}
\begin{tabular}{c|c c c|c c c|c c c}
\hline
\multirow{2}{*}{Method}    &\multicolumn{3}{c|}{3D Detection on Car}  &\multicolumn{3}{c|}{3D Detection on Cyclist} &\multicolumn{3}{c}{3D Detection on Pedestrian}\\
\cline{2-10}
&Mod & Easy &Hard  &Mod & Easy &Hard &Mod & Easy &Hard\\
\hline
SECOND \cite{yan2018second} &$81.96$ &$90.95$ &$77.24$ &$61.62$ &$80.13$ &$57.77$ &$64.19$ &$69.14$ &$57.99$\\
Ours (Dense) & $82.2$ & $91.09$ &  $79.69$ & $66.45$ &$85.85$ &$62.16$ & $62.82$ & $68.88$ &$55.78$\\
Ours ($C=\inf$) & $\textbf{82.93}$ & $\textbf{91.98}$	&$\textbf{80.46}$  & $67.66$ & $86.41$	&$63.5$ & $62.08$ & $68.22$	&$56.64$ \\
Ours ($C=64$)   & $82.75$ & $91.87$	&$80.22$   & $\textbf{72.55}$ & $\textbf{88.22}$	&$\textbf{68.08}$  & $\textbf{65.9}$ & $\textbf{72.23}$	&$\textbf{60.06}$\\
\hline

\end{tabular}
\end{center}
\caption{Effect of different target assignment strategy. \textbf{Dense}: assigning both empty and non-empty voxels inside objects as hotspots; $\textbf{C=inf}$: assigning all spots as hotspots; $\textbf{C=64}$: assigning limited number of spots as hotspots. }\label{ablation-limit}
\end{table*}

\subsection{Ablation Studies} \label{ablation}
\subsubsection{Effect of different target assignment strategies}
We show the effect of our hotspot assignment strategy in Table \ref{ablation-limit}.  We present three types of target assignment strategy for hotspot while keeping all other settings the same. 1) Dense means we assign all voxels (empty and non-empty) inside objects as hotspots while ignoring voxels around ground truth bounding box boundaries. 2) We assign all non-empty spots as hotspots, corresponding to $C=\inf$ in Table \ref{ablation-limit}. The maximum number of hotspots in each object is $M=\frac{C}{Vol}$ as explained in Sec. \ref{hotspot}. 3) We set $C=64$ in our approach to adaptively limit the number of hotspots in each objects. For reference, we also include our baseline, SECOND \cite{yan2018second}. The results show that ours (Dense) and ours ($C=\inf$) have similar performances. When considering pedestrian detection ours ($C=\inf$) is slightly better than ours (Dense). Compared to SECOND, they are both better in car and cyclist detection, especially in the hard cases, but worse in pedestrian detection. The inter-object point-sparsity imbalance makes the pedestrian category hard to train. After balancing the number of hotspots over all objects, ours ($C=64$) outperforms all other target assignment strategies by a large margin in both cyclist and pedestrian detection, while the performance for cars barely changes. This justifies our motivation to force the network to learn the minimal and most discriminative features for each objects. 

\subsubsection{Effect of spatial relation encoder} To prove the effectiveness of our hotspot spatial encoder, we show the results of our HotSpotNet with and without spatial relation encoder on KITTI validation split for cars in Table \ref{alation-part}. We can see that when our algorithm is trained with the spatial relation encoder, the overall performance is boosted. Especially, the great improvement can be observed in hard cases for cyclists and pedestrians. 

\begin{table*}[h]
\begin{center}
\begin{tabular}{c|c c c|c c c|c c c}
\hline
\multirow{2}{*}{Method}    &\multicolumn{3}{c|}{3D Detection on Car}  &\multicolumn{3}{c|}{3D Detection on Cyclist} &\multicolumn{3}{c}{3D Detection on Pedestrian}\\
\cline{2-10}
&Mod & Easy &Hard  &Mod & Easy &Hard &Mod & Easy &Hard\\
\hline
Ours w/o quadrant & $82.27$ & $91.75$	&$79.96$  & $69.31$ & $\textbf{89.48}$	&$65.04$ & $65.45$ & $\textbf{72.77}$	&$58.36$ \\
Ours w quadrant   & $\textbf{82.75}$ & $\textbf{91.87}$	&$\textbf{80.22}$   & $\textbf{72.55}$ & $88.22$	&$\textbf{68.08}$  & $\textbf{65.9}$ & $72.23$	&$\textbf{60.06}$\\
\hline
Diff & $\uparrow \textbf{0.48}$ &$\uparrow \textbf{0.12}$ &$\uparrow \textbf{0.26}$ &$\uparrow \textbf{3.24}$ & $\downarrow -1.24$ &$\uparrow \textbf{3.04}$ &$\uparrow \textbf{0.45}$ &$\downarrow -0.54$ &$\uparrow \textbf{1.7}$\\
\hline

\end{tabular}
\caption{Effect of quadrants as spatial relation encoding.}\label{alation-part}
\end{center}
\end{table*}

\subsubsection{Effect of soft $argmin$} We show the importance of soft $argmin$ in Table \ref{alation-softargmin}. We perceive improvements by using soft $argmin$ instead of the raw values. Particularly on small objects, e.g. cyclists and pedestrians, soft $argmin$ considerably improves the performance by avoiding regression on absolute values with different scales.

\begin{table*}[h]
\begin{center}
\begin{tabular}{c|c c c|c c c|c c c}
\hline
\multirow{2}{*}{Method}    &\multicolumn{3}{c|}{3D Detection on Car} &\multicolumn{3}{c|}{3D Detection on Cyclist} &\multicolumn{3}{c}{3D Detection on Pedestrian}\\
\cline{2-10}
&Mod & Easy &Hard &Mod & Easy &Hard &Mod & Easy &Hard\\
\hline
Ours w/o soft $argmin$ & $82.31$ & $91.53$	&$79.88$ & $68.65$ & $88.11$	&$64.36$ & $63.7$ & $67.62$	&$57.15$  \\
Ours w/ soft $argmin$     & $\textbf{82.75}$ & $\textbf{91.87}$ &$\textbf{80.22}$   & $\textbf{72.55}$ & $\textbf{88.22}$	&$\textbf{68.08}$  & $\textbf{65.9}$ & $\textbf{72.23}$	&$\textbf{60.06}$  \\
\hline
Diff & $\uparrow \textbf{0.44}$ &$\uparrow \textbf{0.34}$ & $\uparrow \textbf{0.34}$ &$\uparrow \textbf{3.9}$& $\uparrow \textbf{0.11}$& $\uparrow \textbf{3.72}$& $\uparrow \textbf{2.9}$ &$\uparrow \textbf{4.59}$ &$\uparrow \textbf{2.91}$\\
\hline
\end{tabular}
\caption{Performance of soft $argmin$ on $(x,y,z)$ coordination.}\label{alation-softargmin}
\end{center}
\end{table*}

\begin{figure*}[h!]
\includegraphics[width=\linewidth]{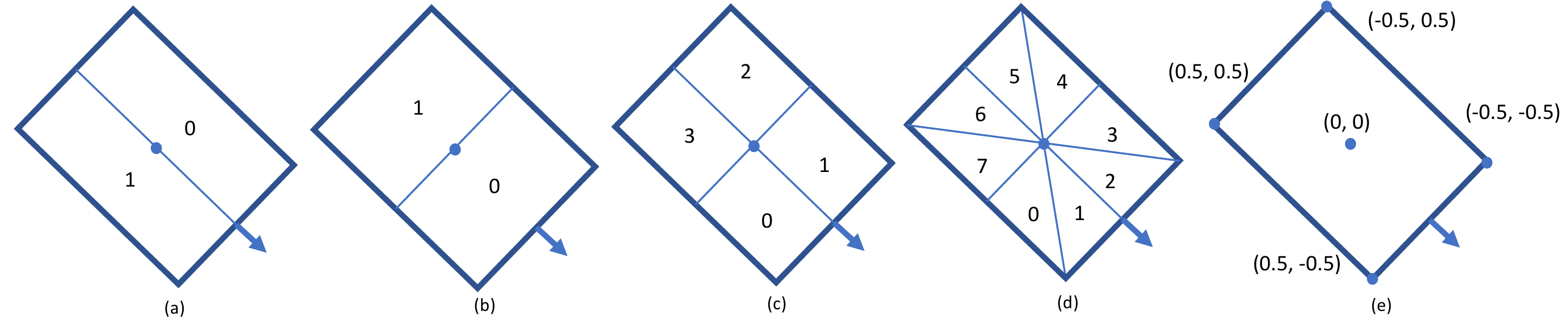}
\caption{Different hotspot-object spatial relation encodings in local object coordinate system. (a) classifying the hotspot location into left or right part of object bounding box. (b) classifying the hotspot location into front or back of object bounding box. (c) classifying the hotspot location into quadrants of object bounding box. (d) classifying the hotspot location into 8 directions of object bounding box. (e) regressing the hotspot location relative to the object center. The object center is the origin. The farther away from the center, the higher the absolute degree of deviation ($0.5$). The values are normalized by the sizes of the bounding box.}
\label{fig:encoding}
\end{figure*}

\subsubsection{Effects of different spatial relation encodings}
\begin{table*}[!h]
\renewcommand{\arraystretch}{1.2}
\begin{center}
\begin{tabular}{c|c c c|c c c|c c c}
\hline
\multirow{2}{*}{Method}    &\multicolumn{3}{c|}{3D Detection on Car}  &\multicolumn{3}{c|}{3D Detection on Cyclist}  &\multicolumn{3}{c}{3D Detection on Pedestrian}\\
\cline{2-10}
&Mod & Easy &Hard &Mod & Easy &Hard &Mod & Easy &Hard\\
\hline
Ours w/o spatial relation &$82.27$ & $91.75$	&$79.96$  & $69.31$ & $\textbf{89.48}$	&$65.04$ & $65.45$ & $\textbf{72.77}$	&$58.36$  \\
Ours w/ left\&right & $82.25$ & $91.62$	&$79.66$  & $69.56$ & $88.42$	&$65.38$  & $64.05$ & $69.58$	&$57.78$\\
Ours w/ front\&back & $82.42$ & $91.88$	&$80.03$  & $69.07$ & $87.51$	&$64.76$ & $65.18$ & $71.45$	&$59.44$ \\
Ours w quadrant   & $\textbf{82.75}$ & $91.87$	&$\textbf{80.22}$   & $\textbf{72.55}$ & $88.22$	&$\textbf{68.08}$  & $65.90$ & $72.23$	&$\textbf{60.06}$\\
Ours w/ 8 directions  & $82.66$ & $\textbf{92.04}$	&$80.21$    & $69.99$	&$86.29$   & $64.94$  & $\textbf{66.26}$ & $71.11$	&$58.92$ \\
Ours w/ deviation regression & $82.03$ &$91.92$ &$79.66$ & $70.25$ &$87.29$ &$65.29$ & $65.13$ &$71.42$ &$58.77$\\
\hline
\multirow{2}{*}{Method}  & \multicolumn{3}{c|}{BEV Detection on Car} & \multicolumn{3}{c|}{BEV Detection on Cyclist} & \multicolumn{3}{c}{BEV Detection on Pedestrian}\\
\cline{2-10}
&Mod & Easy &Hard &Mod & Easy &Hard &Mod & Easy &Hard\\
\hline
Ours w/o spatial relation & $89.29$ & $95.75$	&$\textbf{88.86}$  &$71.63$ &$\textbf{90.63}$ &$67.25$ &$68.86$ &$76.38$ &$62.93$  \\
Ours w/ left\&right & $89.08$ & $95.42$	&$88.80$   & $72.04$ & $90.07$	&$67.91$  & $67.96$ & $73.71$	&$62.45$ \\
Ours w/ front\&back & $89.06$ & $95.60$	&$88.62$  & $71.95$ & $89.28$	&$67.72$ & $70.18$ & $76.42$	&$64.53$\\
Ours w quadrant   & $\textbf{89.67}$ & $\textbf{95.88}$	&$87.23$ & $\textbf{74.97}$ & $90.41$	&$\textbf{70.53}$ & $69.28$ & $75.83$	&$63.58$ \\
Ours w/ 8 directions & $89.22$ & $95.82$	&$88.75$  & $71.11$ & $87.04$	&$66.46$ & $\textbf{70.55}$ & $\textbf{76.46}$	&$\textbf{63.94}$\\
Ours w/ deviation regression & $89.10$ &$95.77$ &$88.58$  & $72.32$ &$88.83$ &$68.04$   & $69.85$ &$75.12$ &$62.80$\\
\hline
\end{tabular}
\caption{Effects of different hotspot spatial relation encodings.}\label{alation-part}
\end{center}
\end{table*}

Besides the quadrant partition presented in the main paper, we present four more types of encodings as shown in Fig. \ref{fig:encoding}. We supervise our network using different spatial encoding targets: 1) classifying hotspot location into left or right part of the object; 2) classifying hotspot location into front or back of the object; 3) classifying hotspot location into quadrants of the objects; 4) classifying hotspot location into eight directions of the objects; 5) directly regressing deviation to the object center. Deviation is two decimals denoting the relative deviations from the center along the box width and length, and ranges within $[-0.5, 0.5]$ because we normalize the values by the box width and length. Thus, $(0, 0)$ is the center of the box and the four corners are $(-0.5, -0.5)$, $(-0.5, 0.5)$, $(0.5, -0.5)$ and $(0.5, 0.5)$. The performance of our approach without any spatial relation encoding is presented using `Ours w/o directions'. The performances of integrating different encodings into our approach are listed in Table \ref{alation-part}. Generally, too coarse, e.g. two partitions, left$\&$right or front$\&$back, or too sophisticated, e.g. eight directions, hotspot-object encoding relations does not help the regression.  By contrast, quadrant partition can improve the performance. We argue that quadrant partition encodes the coarse spatial location of the hotspots which helps the final accurate localization.

\subsection{Qualitative Visualization} \label{ablation}

\begin{figure}[h]
\center
\includegraphics[width=\linewidth, height=7cm]{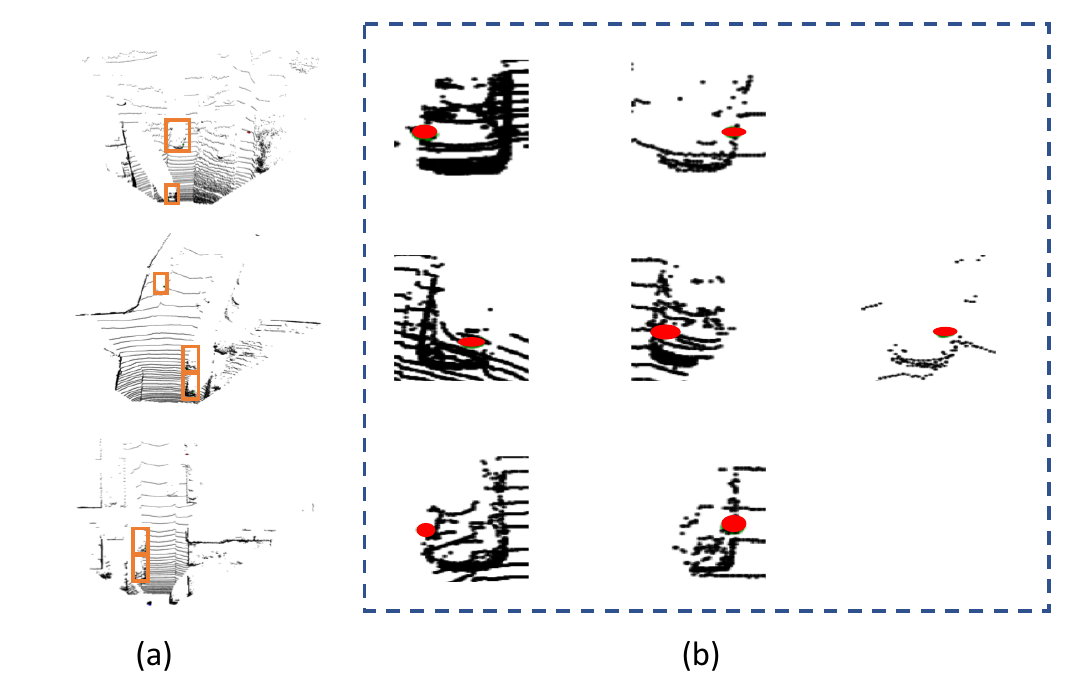}
\caption{Hotspots visualization. (a) is the original LiDAR point clouds visualization in BEV. All the cars in (a) with active hotspots (colored in red) are visualized in (b). Better viewed in color and zoom in. 
}
\label{fig:hotspots}
\end{figure}

Previously, we introduce the concept of hotspots and their assignment methods. Do we really learn the hotspots and what do they look like? We trace our detection bounding boxes results back to original fired voxels and visualize them in Fig. \ref{fig:hotspots}. Here we visualize some samples with cars from validation dataset, all the fired hotspots are marked red. (a) presents the original LiDAR point clouds in BEV and (b) shows all the point clouds from the detected cars. Interestingly, all the fired hotspots sit at the front corner of the car. It shows that the front corner may be the most distinctive `part' for detecting/representing a car.

\section{Conclusion}
We propose a novel representation, Object-as-Hotspots and an anchor-free detection head with its unique target assignment strategy to tackle inter-object point-sparsity imbalance. Spatial relation encoding as quadrants strengthens features of hotspots and further boosts accurate 3D localization. Extensive experiments show that our approach is effective and robust to sparse point clouds. Meanwhile we address regression target imbalance by carefully designing regression targets, among which soft $argmin$ is applied. We believe our work sheds light on rethinking 3D object representations and understanding characteristics of point clouds and corresponding challenges.

\section{Acknowledgement}
We thank Dr. XYZ , Ernest Cheung (Samsung), Gweltaz Lever (Samsung), and Chenxu Luo (Johns Hopkins University and Samsung) for useful discussions that greatly improved the manuscript. 

{
\bibliographystyle{unsrt}
\bibliographystyle{ieee_fullname}
\bibliography{egpaper_final}

\begin{thebibliography}{10}

\bibitem{zhou2018voxelnet}
Yin Zhou and Oncel Tuzel.
\newblock Voxelnet: End-to-end learning for point cloud based 3d object
  detection.
\newblock In {\em CVPR}, 2018.

\bibitem{maturana2015voxnet}
Daniel Maturana and Sebastian Scherer.
\newblock Voxnet: A 3d convolutional neural network for real-time object
  recognition.
\newblock In {\em IROS}, 2015.

\bibitem{yang2018pixor}
Bin Yang, Wenjie Luo, and Raquel Urtasun.
\newblock Pixor: Real-time 3d object detection from point clouds.
\newblock In {\em CVPR}, 2018.

\bibitem{jin2006context}
Ya~Jin and Stuart Geman.
\newblock Context and hierarchy in a probabilistic image model.
\newblock In {\em CVPR}, 2006.

\bibitem{zhu2008}
Long~Leo Zhu, Chenxi Lin, Haoda Huang, Yuanhao Chen, and Alan Yuille.
\newblock Unsupervised structure learning: Hierarchical recursive composition,
  suspicious coincidence and competitive exclusion.
\newblock In {\em ECCV}, 2008.

\bibitem{fidler2014}
Sanja Fidler, Marko Boben, and Ales Leonardis.
\newblock Learning a hierarchical compositional shape vocabulary for
  multi-class object representation.
\newblock {\em arXiv preprint arXiv:1408.5516}, 2014.

\bibitem{dai2014unsupervised}
Jifeng Dai, Yi~Hong, Wenze Hu, Song-Chun Zhu, and Ying Nian~Wu.
\newblock Unsupervised learning of dictionaries of hierarchical compositional
  models.
\newblock In {\em CVPR}, 2014.

\bibitem{kortylewski2017greedy}
Adam Kortylewski, Aleksander Wieczorek, Mario Wieser, Clemens Blumer, Sonali
  Parbhoo, Andreas Morel-Forster, Volker Roth, and Thomas Vetter.
\newblock Greedy structure learning of hierarchical compositional models.
\newblock {\em arXiv preprint arXiv:1701.06171}, 2017.

\bibitem{zhang2018deepvoting}
Zhishuai Zhang, Cihang Xie, Jianyu Wang, Lingxi Xie, and Alan~L Yuille.
\newblock Deepvoting: A robust and explainable deep network for semantic part
  detection under partial occlusion.
\newblock In {\em CVPR}, 2018.

\bibitem{kendall2017end}
Alex Kendall, Hayk Martirosyan, Saumitro Dasgupta, Peter Henry, Ryan Kennedy,
  Abraham Bachrach, and Adam Bry.
\newblock End-to-end learning of geometry and context for deep stereo
  regression.
\newblock In {\em ICCV}, 2017.

\bibitem{zhou2019bottom}
Xingyi Zhou, Jiacheng Zhuo, and Philipp Krahenbuhl.
\newblock Bottom-up object detection by grouping extreme and center points.
\newblock In {\em CVPR}, pages 850--859, 2019.

\bibitem{law2018cornernet}
Hei Law and Jia Deng.
\newblock Cornernet: Detecting objects as paired keypoints.
\newblock In {\em ECCV}, pages 734--750, 2018.

\bibitem{zhou2019objects}
Xingyi Zhou, Dequan Wang, and Philipp Kr{\"a}henb{\"u}hl.
\newblock Objects as points.
\newblock {\em arXiv preprint arXiv:1904.07850}, 2019.

\bibitem{duan2019centernet}
Kaiwen Duan, Song Bai, Lingxi Xie, Honggang Qi, Qingming Huang, and Qi~Tian.
\newblock Centernet: Keypoint triplets for object detection.
\newblock In {\em ICCV}, pages 6569--6578, 2019.

\bibitem{tian2019fcos}
Zhi Tian, Chunhua Shen, Hao Chen, and Tong He.
\newblock Fcos: Fully convolutional one-stage object detection.
\newblock {\em arXiv preprint arXiv:1904.01355}, 2019.

\bibitem{zhu2019feature}
Chenchen Zhu, Yihui He, and Marios Savvides.
\newblock Feature selective anchor-free module for single-shot object
  detection.
\newblock {\em arXiv preprint arXiv:1903.00621}, 2019.

\bibitem{geirhos2018imagenettrained}
Robert Geirhos, Patricia Rubisch, Claudio Michaelis, Matthias Bethge, Felix~A.
  Wichmann, and Wieland Brendel.
\newblock Imagenet-trained {CNN}s are biased towards texture; increasing shape
  bias improves accuracy and robustness.
\newblock In {\em International Conference on Learning Representations}, 2019.

\bibitem{wang2018sgpn}
Weiyue Wang, Ronald Yu, Qiangui Huang, and Ulrich Neumann.
\newblock Sgpn: Similarity group proposal network for 3d point cloud instance
  segmentation.
\newblock In {\em CVPR}, 2018.

\bibitem{yang2019learning}
Bo~Yang, Jianan Wang, Ronald Clark, Qingyong Hu, Sen Wang, Andrew Markham, and
  Niki Trigoni.
\newblock Learning object bounding boxes for 3d instance segmentation on point
  clouds.
\newblock {\em arXiv preprint arXiv:1906.01140}, 2019.

\bibitem{meyer2019lasernet}
Gregory~P Meyer, Ankit Laddha, Eric Kee, Carlos Vallespi-Gonzalez, and Carl~K
  Wellington.
\newblock Lasernet: An efficient probabilistic 3d object detector for
  autonomous driving.
\newblock In {\em CVPR}, 2019.

\bibitem{qi2019deep}
Charles~R Qi, Or~Litany, Kaiming He, and Leonidas~J Guibas.
\newblock Deep hough voting for 3d object detection in point clouds.
\newblock In {\em ICCV}, 2019.

\bibitem{shi2019pointrcnn}
Shaoshuai Shi, Xiaogang Wang, and Hongsheng Li.
\newblock Pointrcnn: 3d object proposal generation and detection from point
  cloud.
\newblock In {\em CVPR}, 2019.

\bibitem{kong2019foveabox}
Tao Kong, Fuchun Sun, Huaping Liu, Yuning Jiang, and Jianbo Shi.
\newblock Foveabox: Beyond anchor-based object detector.
\newblock {\em arXiv preprint arXiv:1904.03797}, 2019.

\bibitem{lin2017feature}
Tsung-Yi Lin, Piotr Doll{\'a}r, Ross Girshick, Kaiming He, Bharath Hariharan,
  and Serge Belongie.
\newblock Feature pyramid networks for object detection.
\newblock In {\em CVPR}, 2017.

\bibitem{ren2015faster}
Shaoqing Ren, Kaiming He, Ross Girshick, and Jian Sun.
\newblock Faster r-cnn: Towards real-time object detection with region proposal
  networks.
\newblock In {\em Neural Information Processing Systems}, 2015.

\bibitem{girshick2015fast}
Ross Girshick.
\newblock Fast r-cnn.
\newblock In {\em ICCV}, 2015.

\bibitem{lin2017focal}
Tsung-Yi Lin, Priya Goyal, Ross Girshick, Kaiming He, and Piotr Doll{\'a}r.
\newblock Focal loss for dense object detection.
\newblock In {\em ICCV}, 2017.

\bibitem{liu2016ssd}
Wei Liu, Dragomir Anguelov, Dumitru Erhan, Christian Szegedy, Scott Reed,
  Cheng-Yang Fu, and Alexander~C Berg.
\newblock Ssd: Single shot multibox detector.
\newblock In {\em ECCV}, pages 21--37. Springer, 2016.

\bibitem{geiger2012we}
Andreas Geiger, Philip Lenz, and Raquel Urtasun.
\newblock Are we ready for autonomous driving? the kitti vision benchmark
  suite.
\newblock In {\em CVPR}, 2012.

\bibitem{caesar2019nuscenes}
Holger Caesar, Varun Bankiti, Alex~H Lang, Sourabh Vora, Venice~Erin Liong,
  Qiang Xu, Anush Krishnan, Yu~Pan, Giancarlo Baldan, and Oscar Beijbom.
\newblock nuscenes: A multimodal dataset for autonomous driving.
\newblock {\em arXiv preprint arXiv:1903.11027}, 2019.

\bibitem{yan2018second}
Yan Yan, Yuxing Mao, and Bo~Li.
\newblock Second: Sparsely embedded convolutional detection.
\newblock {\em Sensors}, 18(10):3337, 2018.

\bibitem{zhu2019class}
Benjin Zhu, Zhengkai Jiang, Xiangxin Zhou, Zeming Li, and Gang Yu.
\newblock Class-balanced grouping and sampling for point cloud 3d object
  detection.
\newblock {\em arXiv preprint arXiv:1908.09492}, 2019.

\bibitem{he2016deep}
Kaiming He, Xiangyu Zhang, Shaoqing Ren, and Jian Sun.
\newblock Deep residual learning for image recognition.
\newblock In {\em CVPR}, pages 770--778, 2016.

\bibitem{lang2019pointpillars}
Alex~H Lang, Sourabh Vora, Holger Caesar, Lubing Zhou, Jiong Yang, and Oscar
  Beijbom.
\newblock Pointpillars: Fast encoders for object detection from point clouds.
\newblock In {\em CVPR}, 2019.

\bibitem{loshchilov2017fixing}
Ilya Loshchilov and Frank Hutter.
\newblock Fixing weight decay regularization in adam.
\newblock {\em arXiv preprint arXiv:1711.05101}, 2017.

\bibitem{smith2019super}
Leslie~N Smith and Nicholay Topin.
\newblock Super-convergence: Very fast training of neural networks using large
  learning rates.
\newblock In {\em Artificial Intelligence and Machine Learning for Multi-Domain
  Operations Applications}, volume 11006, page 1100612. International Society
  for Optics and Photonics, 2019.

\bibitem{simon2018complex}
Martin Simon, Stefan Milz, Karl Amende, and Horst-Michael Gross.
\newblock Complex-yolo: An euler-region-proposal for real-time 3d object
  detection on point clouds.
\newblock In {\em ECCV}, 2018.

\bibitem{zhou2019iou}
Dingfu Zhou, Jin Fang, Xibin Song, Chenye Guan, Junbo Yin, Yuchao Dai, and
  Ruigang Yang.
\newblock Iou loss for 2d/3d object detection.
\newblock {\em arXiv preprint arXiv:1908.03851}, 2019.

\bibitem{wang2019voxel}
Bei Wang, Jianping An, and Jiayan Cao.
\newblock Voxel-fpn: multi-scale voxel feature aggregation in 3d object
  detection from point clouds.
\newblock {\em arXiv preprint arXiv:1907.05286}, 2019.

\bibitem{liang2018deep}
Ming Liang, Bin Yang, Shenlong Wang, and Raquel Urtasun.
\newblock Deep continuous fusion for multi-sensor 3d object detection.
\newblock In {\em ECCV}, 2018.

\bibitem{chen2017multi}
Xiaozhi Chen, Huimin Ma, Ji~Wan, Bo~Li, and Tian Xia.
\newblock Multi-view 3d object detection network for autonomous driving.
\newblock In {\em CVPR}, 2017.

\bibitem{ku2018joint}
Jason Ku, Melissa Mozifian, Jungwook Lee, Ali Harakeh, and Steven~L Waslander.
\newblock Joint 3d proposal generation and object detection from view
  aggregation.
\newblock In {\em IROS}, 2018.

\bibitem{qi2018frustum}
Charles~R Qi, Wei Liu, Chenxia Wu, Hao Su, and Leonidas~J Guibas.
\newblock Frustum pointnets for 3d object detection from rgb-d data.
\newblock In {\em CVPR}, 2018.

\bibitem{wang2019frustum}
Zhixin Wang and Kui Jia.
\newblock Frustum convnet: Sliding frustums to aggregate local point-wise
  features for amodal 3d object detection.
\newblock In {\em IROS}, 2019.

\bibitem{liang2019multi}
Ming Liang, Bin Yang, Yun Chen, Rui Hu, and Raquel Urtasun.
\newblock Multi-task multi-sensor fusion for 3d object detection.
\newblock In {\em CVPR}, 2019.

\bibitem{Chen_2019_ICCV}
Yilun Chen, Shu Liu, Xiaoyong Shen, and Jiaya Jia.
\newblock Fast point r-cnn.
\newblock In {\em ICCV}, October 2019.

\bibitem{yang2019std}
Zetong Yang, Yanan Sun, Shu Liu, Xiaoyong Shen, and Jiaya Jia.
\newblock Std: Sparse-to-dense 3d object detector for point cloud.
\newblock {\em arXiv preprint arXiv:1907.10471}, 2019.

\bibitem{yin2020center}
Tianwei Yin, Xingyi Zhou, and Philipp Kr{\"a}henb{\"u}hl.
\newblock Center-based 3d object detection and tracking.
\newblock {\em arXiv:2006.11275}, 2020.

\bibitem{ye2020sarpnet}
Yangyang Ye, Houjin Chen, Chi Zhang, Xiaoli Hao, and Zhaoxiang Zhang.
\newblock Sarpnet: Shape attention regional proposal network for lidar-based 3d
  object detection.
\newblock {\em Neurocomputing}, 379:53--63, 2020.

\bibitem{hu2019you}
Peiyun Hu, Jason Ziglar, David Held, and Deva Ramanan.
\newblock What you see is what you get: Exploiting visibility for 3d object
  detection.
\newblock {\em arXiv preprint arXiv:1912.04986}, 2019.

\end{thebibliography}
}

\end{document}